# Word Searching in Scene Image and Video Frame in Multi-Script Scenario using Dynamic Shape Coding

[a]Partha Pratim Roy*, [b]Ayan Kumar Bhunia, [b]Avirup Bhattacharyya [c]Umapada Pal

[a]Dept. of CSE, Indian Institute of Technology Roorkee, India
[b]Dept. of ECE, Institute of Engineering & Management, Kolkata, India
[c]CVPR Unit, Indian Statistical Institute, Kolkata, India.
[a]email: proy.fcs@iitr.ac.in, TEL: +91-1332-284816

## Abstract

Retrieval of text information from natural scene images and video frames is a challenging task due to its inherent problems like complex character shapes, low resolution, background noise, etc. Available OCR systems often fail to retrieve such information in scene/video frames. Keyword spotting, an alternative way to retrieve information, performs efficient text searching in such scenarios. However, current word spotting techniques in scene/video images are script-specific and they are mainly developed for Latin script. This paper presents a novel word spotting framework using dynamic shape coding for text retrieval in natural scene image and video frames. The framework is designed to search query keyword from multiple scripts with the help of on-the-fly script-wise keyword generation for the corresponding script. We have used a two-stage word spotting approach using Hidden Markov Model (HMM) to detect the translated keyword in a given text line by identifying the script of the line. A novel unsupervised dynamic shape coding based scheme has been used to group similar shape characters to avoid confusion and to improve text alignment. Next, the hypotheses locations are verified to improve retrieval performance. To evaluate the proposed system for searching keyword from natural scene image and video frames, we have considered two popular Indic scripts such as Bangla (Bengali) and Devanagari along with English. Inspired by the zone-wise recognition approach in Indic scripts[1], zone-wise text information has been used to improve the traditional word spotting performance in Indic scripts. For our experiment, a dataset consisting of images of different scenes and video frames of English, Bangla and Devanagari scripts were considered. The results obtained showed the effectiveness of our proposed word spotting approach.

*Keywords-* Scene and Video text retrieval, Indic word spotting, Hidden Markov Model, Dynamic shape code, Word spotting in multiple scripts.

# 1. Introduction

With the rapid progress of internet and mobile technology, there is a large amount of digital information recorded every day in form of videos and scene images. These images/videos are being uploaded to social network sites from every corner of the world. An efficient indexing approach is thus necessary to retrieve the required information from this large dataset. If text is available in these images/videos, such textual information can be useful in automatic video indexing and retrieval. Querying with text information can be effective to search required image/video images[2]. In multilingual and multi-script countries like India, information communication using multiple languages/scripts is quite common. For example, people belonging to rural or semi-urban areas may not be as proficient in English as their local dialect. In order meet this end, various social networking sites and smartphones are coming up with features for communication in regional languages apart from English. Hence, a query word from the user can appear in different scripts in these multi-lingual images/videos. Thus, a word-spotting system [3]–[5] that can retrieve scene/video images from different scripts can be very useful. To the best of our knowledge, searching of scene/video text in multi-lingual scripts is hardly reported in the literature. Existing text searching approaches generally consider query and target text in a single language to search images/videos. Thus, images/videos containing same query word in different scripts cannot be retrieved with such systems.

Available character recognition systems[1] which were developed mainly for scanned document images do not work properly in scene/video images because of their poor quality. The major difficulties in recognizing the in-video text are low-resolution, noisy background image, immense variations in color, etc. Hence word searching using a recognition system in such images is a difficult task. The alternative approach "word spotting" by which similar words are searched based on query image or query keyword is more effective [6]–[9]. It is because word-spotting techniques are developed to retrieve words by looking the patterns in the images and thus it performs better in scenarios where recognition is not always easy. Though there exists a number of approaches for text spotting in Latin text [10], [11], not many approaches have focussed on other scripts, especially Indic scripts, for word spotting techniques in image/video

---

[1] https://code.google.com/p/tesseract-ocr/

frames [12]. None of these existing approaches described any multilingual word spotting scenarios.

Also, the word-spotting approaches available in literature are mainly developed to search keywords in script specific way. Most of the word-spotting techniques in scene image/video frames search keywords in English text. These techniques cannot be used to search query keywords in images/videos of different scripts. It would be very effective if images/video frames of different scripts can be searched from an archive with a single query word. It will be a step forward to the existing keyword searching approaches. A keyword of a particular script can be translated to different scripts and those generated keywords may be used for searching different scripts in the archive. One of the main problems which hinder this process is modeling a large number of characters of target scripts. When the target script is Indic, the complexity is more [1]. Characters in Indic scripts appear with complex syntax and spatial variation of the characters when combined with other characters to form a word. Word spotting of these scripts needs special care due to several reasons; a) complex shape of Indic script characters, b) less similarity among inter-character classes, c) presence of characters in three zones unlike English text, etc. There exist only a few works on Indic text recognition in natural scene images/video frames [13]. The authors segmented the characters explicitly from the words and next character recognition was performed. Word level recognition was not reported.

Recently researchers are developing system for word segmentation before searching [10]. However, these approaches may fail when the spaces between characters and words in a text line are not uniform. Traditional HMM-based word spotting approaches use keyword-specific filler model in a single stage which overcomes the cursive and complex shapes of characters and words but they may not work to provide the accurate location of the query keyword in a text line image. This is mainly due to large number of character classes and its variation. Due to appearance of some similar character shapes in two different words, the traditional word spotting approaches provide high score which reduces the precision and recall performance. Also, proper boundary detection of query keyword is important in word spotting. It happens sometimes when the characters of neighbour words and query keyword share some similar characters. While searching for the name of a book, city, or a person's name, the results are not satisfying if the boundaries of the target words are not marked properly. To avoid this drawback, in this paper we propose a two-stage word spotting framework for scene/video images. First, similar character shapes are grouped together in a filler model of word spotting to improve locating similar word shapes in text lines and improve the word alignment. The word location hypotheses are next verified using actual character set.

This paper presents a novel word spotting framework using dynamic shape coding for text retrieval in natural scene image and video frames. The framework has been designed to search a query keyword from different scripts with the help of on-the-fly script-wise query keyword translation and searching in target script. After identifying the script of target text line, our proposed word spotting approach consists of two steps using Hidden Markov Model (HMM). In the first step, we detect a coarse-level word location for the query keyword using a reduced character-set. The reduced character set is obtained using dynamic shape coding which in turn improves recall performance. Next, that particular location of the text line image is verified with actual character-set of corresponding script. Thus improves the precision of word detection performance. The two-step word-spotting approach uses contextual information from the neighbour to enrich the shape feature. To our knowledge dynamic shape coding has not been used for word spotting in earlier research work. The proposed approach improves the existing systems in two ways. Firstly, due to the nature of shape coding scheme, it provides more accurate location of the query keyword with minimum overlapping with the neighbouring words. Secondly, due to training of HMM model from segmented word images in verification stage, it improved average retrieval performance.

Some examples of line images of scene images and video frames containing Indic text information are retrieved against the query keyword "College" (see Fig.1). The line images in multiple scripts namely Latin, Devanagari and Bengali are found where the query word is marked in red rectangular box. Unlike Latin, character-modifiers of Bengali, Devanagari, and some other scripts are attached to the consonant (appearing only in middle zone) at any of the 3 zones- upper, middle or lower. Inspired by the zone-wise recognition approach in Indic scripts [1], [14] we consider here zone segmentation approach in scene/video text and use the zone-wise information to improve the traditional word spotting performance.

Fig.1: Example showing word spotting in three different scripts (a). English, (b) Devanagari and (c) Bangla for the query keyword 'College'

The contributions of this paper are three-folds. Firstly, we present a keyword spotting approach that takes a query word and searches the text in multiple scripts from scene image/video frames. Secondly, a two-stage word spotting approach is proposed to improve the traditional word spotting methods. In first step, dynamic shape coding is introduced to group similar shape characters in single class in unsupervised way to provide better performance and word alignment. Dynamic shape coding scheme which does not require any prior knowledge of script for character grouping has not been used earlier for word spotting purpose. In the next step the words in hypotheses locations are verified. Thirdly, the framework has been tested in the multi-script environment to demonstrate the robustness of the proposed system. To the best of our knowledge, word spotting has not been explored in scene image and video frames of Indic scripts.

The rest of the paper is organized as follows. We will present the related work in Section 2. Section 3 describes the overall framework and provides a detailed description of the different steps involved in the framework like enhancement of text line images, binarization, script identification, keyword generation, keyword translation and dynamic shape coding scheme. Section 4 shows the experiments and the detailed results that have been performed. Finally, conclusion and future works are presented in Section 5.

## 2. Related Work

There exist many work on scene/videos for different tasks like visual tracking [62, 63, 64], segmentation [65]. A number of word spotting methods in handwritten/printed [10]-[11], [15], [67] and historical documents[7], [16]–[18], have been proposed in the literature. However, there are very few works for text spotting [19], [20]in scene and video images.

Most of the text related works in the scene and video image concern detection and recognition. Generally, there are three steps in order to recognize text in the scene and video [21], mainly text detection, text binarization and text recognition. Detection step detects the presence of text in the frame. Binarization step aims at segregating text pixels from background pixels and finally, text is recognized as human understandable language. The recognition is divided into two kinds of approaches. The first one is based on segmentation and consists of splitting the text word image into isolated character images. The second one is segmentation free method.. The character segmentation task is challenging because of the presence of noise, complex background, and variation in illumination in text image and on the other hand a part of a character may be misclassified with another character due to over segmentation and under segmentation. Therefore significant information might be lost due to above factors which may lead to poor recognition rate. For these reasons, segmentation free approach, which does not involve any character segmentation, becomes more popular. This kind of approach is suitable for retrieving information in scene and video images where we can expect blur, noisy, complex and low-resolution text.

**Text Recognition in Scene and Video Images:** A survey of recent developments in scene text detection and recognition can be found in [21]. Wang et al. [22] proposed lexicon based scene text recognition at word level using Histogram of Oriented Gradients (HOG) features. Maximally Stable Extremal Region (MSER) [23] has been used for uniform connected component region detection for the purpose of scene text detection. An end-to-end system for text recognition has been developed using convolutional neural networks (Wang, Wu, Coates, & Ng, 2012). This allows a common framework to train highly accurate text detector and character recognizer modules. An end-to-end text recognition system has been developed by [22], [25], [26]. The work in [27], [28] used color channel enhanced contrasting extremal region and neural networks for text recognition tasks.From a color natural scene image, six component-trees have been built from its grayscale image, hue and saturation channel images in a perception-based illumination-invariant color space, and their inverted images, respectively. From each such component-tree, color-enhanced CERs are extracted as character candidates. The algorithm in [29], [30] is different in the sense that it exploits the symmetry property of character groups and allows for direct extraction of text lines from natural images. The method adopted in [20], [31] presents an end-to-end system for text spotting, localizing and recognizing text in natural scene images and text based image retrieval. The method is based on a region proposal mechanism for detection and uses deep convolutional neural networks for recognition. The researchers in [32] used a Deep-Text recurrent network that considers text reading as a sequence labeling problem.The method in [31] used a unified distance metric learning framework for adaptive hierarchical clustering, which can simultaneously learn similarity weights and the clustering threshold to automatically determine the number of clusters and an effective multi-orientation scene text detection system, which constructs text candidates by grouping characters based on this adaptive clustering.

Though there exist a number of works [33]–[36] on scene text detection, only a few methods are reported for video text recognition [25], [37], [38] due to its inherent challenges in video text. An automatic binarization method for color text in video images is proposed using convolutional network [37]. In this approach, a large amount of data is required in training due to the classifier properties. In the same way, Zhou et al. [39] proposed a Canny-based edge based text binarization approach to improve text recognition performance. It used flood fill algorithm to remove the gap on contour thus completes the character contour. Due to this nature, the method is sensitive to seed points.

**Word Spotting in Scene and Video Images:** As we discussed, there are a few works dealing with word spotting in the scene and video images. For example, Jaderberg et al. [40] proposed a segmentation based word spotting method in scene image using Convolution Neural Network (CNN). They have modified the existing CNN's architecture by avoiding down sampling for a per-pixel sliding window, and including multi-mode learning with a mixture of linear models. A segmentation free word spotting approach was introduced in the video by Shivakumara et al. [19]. In this work, spotting is based on Texture-Spatial-Features (TSF). First, the set of texture features is applied for identifying text candidates in a word image using k-means clustering. Then, proximity relation among text candidates is found out and spatial arrangement of pixels is analyzed. An advantage of the method is that it is independent of font, font size and robust some extent to distortion. A number of works have been developed recently for text detection in scene images [41], [42] but query word detection in scene/video frames is still not solved in general. In early days of word spotting, query by example (QBE) performed by image template matching [7] was adopted by many researchers. The second technique, query-by-text based word spotting approach [10] usually outperforms the former one. In these approaches, character/word segmentation tasks are avoided by supervised learning model like HMM, BLSTM [11]. These sequential classifiers are trained for each character and occurrence of specific character sequence is determined based on the probability scores. Bidirectional Long Short-Term Memory (BLSTM) hidden nodes and Connectionist Temporal Classification (CTC) output layer has also been explored for Latin keyword spotting in [18]. Recently, a color channel selection based feature extraction method in scene/video images has been studied in [43] to avoid complex binarization method for word recognition. In recent years, many deep learning based word spotting architectures [59-61] have appeared in the literature. Sudholt et al. [59] used a Convolutional Neural Network (CNN) architecture to estimates the PHOC representation of the given word image. PHOCNet [59] achieved promising performance in handwritten keyword spotting. One of major advantages of Ctrl-F-Net [61] is that it does not require word level segmentation for word spotting; it can perform the word spotting at a page level without any word or line level segmentation. Moreover, it can be trained in an end to end manner. In spite of having superior performance over the traditional machine learning based approaches, the major limitation of deep learning based approach is the non-availability of large dataset in many cases. Hence, in such cases like Indic script where large dataset is not available, traditional method is expected to perform better than the deep-model.

**Indic Word Spotting:** Very few works have been done for keyword spotting in Indic script [25], [44]–[46]. Shape code based word-matching technique was used in Indic printed text lines [45]. Here, vertical shaped based feature, zonal information of extreme points, loop shape and position, crossing count and some background information has been used to search a query word. A Segmentation-free method for spotting query word images in Bangla handwritten documents is done in [46] using Heat Kernel signature (HKS) to represent the local characteristics of detected key points. The method in [45] was designed for printed document images and [46] for handwritten scanned documents. However, word spotting in video/scene images is more challenging due to blurred images, complex character shapes, low resolution, and background noise.

## 3. Proposed framework

In this paper, a two stage word-spotting approach is proposed for keyword detection in video frame/scene image of multiple scripts. For this purpose, text lines segmented from the scene image/video frames are binarized using an efficient Bayesian classifier based binarization approach [38]. Next, the script of the text image is identified using HMM. After identifying the script, the query keyword is translated and corresponding characters of that script are used for word spotting. Next, sliding window based features are extracted from text image along with contextual information for word-spotting. To improve the performance, a dynamic shape coding based approach is used during word spotting to combine the similar shape characters in same class. A flowchart of our proposed framework is shown in Fig.2. Details of these modules are discussed in following subsections. In Section 3.1, the pre-processing tasks such as text enhancement and Binarization processes are detailed. Three domain information from RGB, wavelet, and Gradient are integrated in a Bayesian classifier framework to obtain the binary text image. Section 3.2 discusses the script identification and keyword generation methods. Identification of script from a text image is performed using PHOG feature extraction and HMM based script classification. Once the script is identified the query keyword is translated to corresponding script for searching. Finally in Section 3.3, dynamic shape coding based word spotting approach is presented. This is performed in two stages; first stage for hypotheses generation and second stage for verification.

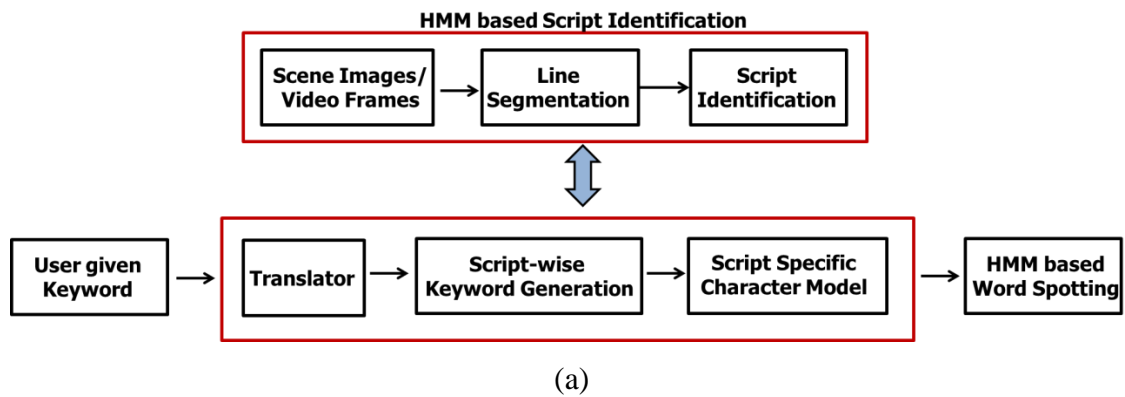

(a)

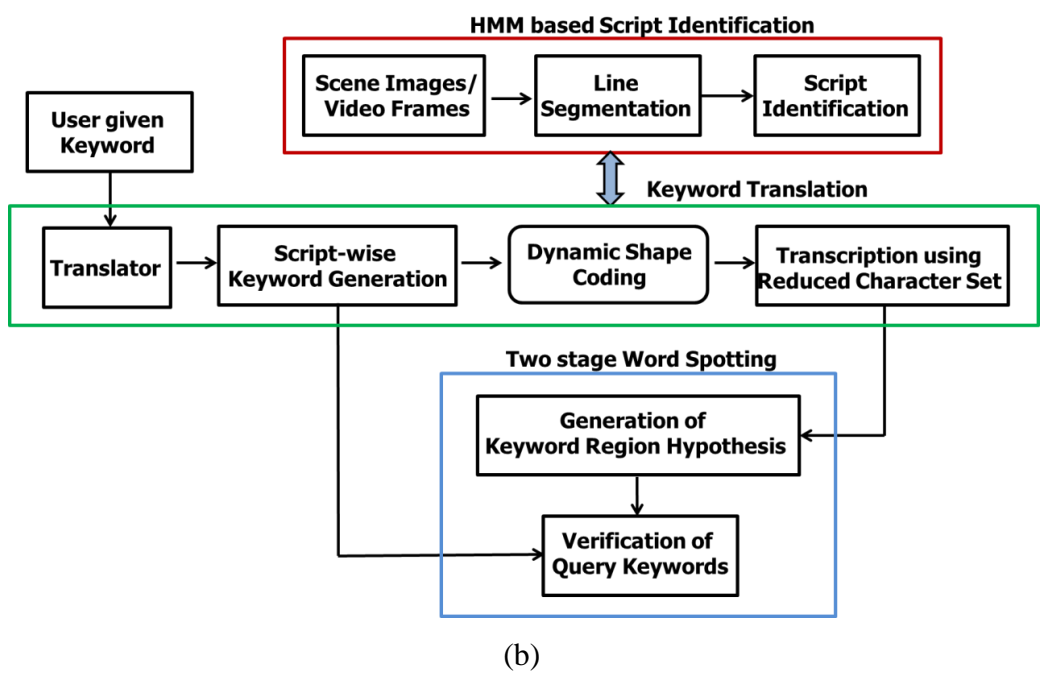

(b)

**Fig.2: (a)** Flowchart of traditional HMM-filler model based keyword spotting paradigm. **(b)** Flowchart of the proposed word-spotting framework. Given a query image and the scene/video frame, the script of the scene/video is identified (marked in red color), next the query word is translated into reduced character set which help in two stage word spotting (marked in blue color).

## 3.1. Enhancement of Text-Line Images and Binarization

In our work, we considered segmented text lines from scene text image or video text frame as an input to our framework. There exists a number of works [27], [29], [36], [47], [68] for text localization and segmentation. In this framework we considered the approach [27] to get the segmented text lines images for our experiment. To enhance the word image in scene image/video for recognition Roy et al. [48] proposed integration of three domains (RGB, wavelet, and Gradient). It was noted that the pixel value of text component might be low in one sub-band but high in another sub-band. The integration of sub-bands of different domains exploits this information for enhancing text in the word image. In our approach, we have used a combination of RGB, wavelet and Gradient-based enhancement approach [48] as pre-processing to obtain an efficient binary image.

To enhance the text pixels in gray image, for each line image, the method decomposes it into R, G, B sub-bands in the color domain, LH (Horizontal), HL (Vertical), HH (Diagonal) in the wavelet domain and Horizontal, Vertical, Diagonal in the gradient domain. Then for each set of sub-bands, three sub-bands are combined to obtain three combined images of the respective domains, namely, RGB-L, Wavelet-L and Gradient-L to improve the fine details at the edge pixel. The linear combination combines three pixels in the respective three sub-band images by adding the three values of each corresponding pixel.

**Text Binarization using Bayesian Classifier:** It was noted that the probability of pixels in RGB-smooth, wavelet-smooth and gradient-smooth being classified as text has high values compared to non-text pixels. Due to this property, a Bayesian classifier for binarization was proposed in [38]. In Bayesian framework, the number of classified text pixels and number of classified non-text pixels are considered as a priori probability of text pixel class and non-text pixel class, which are denoted as P(CTC) and P(NCTC), respectively. P(f(x, y)|TC) denotes the conditional probability of a pixel (x, y) for a given Text Class (TC) which is average of RGB-Smooth, Wavelet-Smooth and Gradient-Smooth and P(f(x, y)|NTC) denotes Non-Text Class (NTC) which is obtained by taking average of complement of RGB-Smooth, Wavelet-Smooth and Gradient-Smooth images. Hence, the conditional probabilities and priori probabilities are substituted in Bayesian framework as given below to obtain posterior probability matrix [38]

$$P(TC|f(x,y)) = \frac{P(f(x,y)|\text{TC})*\text{P(CTC)}}{P(f(x,y)|\text{TC})*\text{P(CTC)}+ P(f(x,y)|\text{NTC})*\text{P(NCTC)}} \ldots\ldots\ldots (1)$$

Then the final binary image (B(x, y)) is obtained with the condition given in following equation on posterior probability matrix.

$$B(x,y) = \begin{cases} 1 & \text{if } P(TC|f(x,y)) \geq \gamma; \\ 0 & \text{Otherwise.} \end{cases} \ldots\ldots\ldots (2)$$

where γ is the threshold parameter which is set to 0.05 [38]. An example of text line and its corresponding binary image are shown in Fig.3.

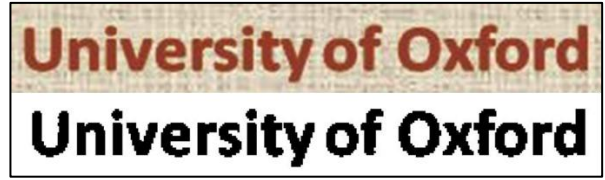
**Fig.3: Binary image using text binarization method** [38]

## 3.2. Script Identification and Script-wise Keyword Generation

This section discusses the identification of script from a text line and then the keyword generation corresponding to identified script. There exist many pieces of work on script identification where the classification is performed at text line, word, or character level [15]. Hidden Markov Model based classifier has been applied successfully for line/word level script [16], [49]. Inspired with this success, our system follows similar HMM-based script identification in scene images and video frames. PHOG (Pyramid Histogram of Oriented Gradients) features[15], described below, are extracted in each sliding window and the feature sequence is fed to HMM classification. The sliding window based feature extraction is applied after normalizing the text height. Block diagram of our script identification system is shown in Fig.4.

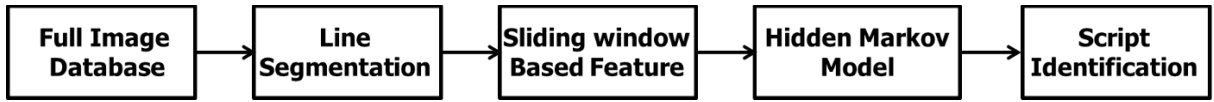

**Fig.4. The block-diagram of the script identification scheme in our system**

**Pyramid Histogram of Oriented Gradient (PHOG) feature:** PHOG[14] is the spatial shape descriptor which gives the feature of the image by spatial layout and local shape, comprising of gradient orientation at each pyramid resolution level. To extract the feature from each sliding window, we have divided it into cells at several pyramid level. The grid has $4^N$ individual cells at *N* resolution level (i.e. *N*=0, 1, 2..). Histogram of gradient orientation of each pixel is calculated from these individual cells and is quantized into *L* bins. Each bin indicates a particular octant in the angular radian space. The concatenation of all feature vectors at each pyramid resolution level provides the final PHOG descriptor. L-vector at level zero represents the L-bins of the histogram at that level. At any individual level, it has $L \times 4^N$ dimensional feature vector where N is the pyramid resolution levels (i.e. N=0, 1, 2….). So, the final PHOG descriptor consists of $L \times \sum_{N=0}^{N=K} 4^N$ dimensional feature vector, where *K* is the limiting pyramid level. In our implementation, we have limited the level (N) to 2 and we considered 8 bins (360º/45º) of angular information. So we obtained (1×8) + (4×8) + (16×8) = (8+32+128) = 168 dimensional feature vector for individual sliding window position.

**HMM-based Script Identification:** We extracted sliding window feature and apply HMM for script identification. The feature vector sequence is processed using left-to-right continuous density HMMs [50]. One of the important features of HMM is the capability to model sequential dependencies. In the past decades, Hidden Markov Models (HMM) has been considered as one of the powerful stochastic approaches. HMM characterizes the temporal observation data that can be discretely or continuously distributed. It has been used successively for modelling sequential data. An HMM can be defined by initial state probabilities π, state transition matrix A =[$a_{ij}$], i, j=1,2,…,N, where $a_{ij}$ denotes the transition probability from state i to state j and output probability $b_j(O_K)$ modeled with continuous output probability density function. The density function is written as $b_j(x)$, where x represents *k* dimensional feature vector. A separate Gaussian mixture model (GMM) is defined for each state of the model. Formally, the output probability density of state *j* is defined as

$$b_j(x) = \sum_{k=1}^{M_j} c_{jk} \mathcal{N}(x, \mu_{jk}, \Sigma_{jk}) \quad \ldots\ldots\ldots (3)$$

where, $M_j$ is the number of Gaussians assigned to *j*. and $\mathcal{N}(x, \mu, \Sigma)$ denotes a Gaussian with mean $\mu$ and covariance matrix $\Sigma$ and $c_{jk}$ is the weight coefficient of the Gaussian component k of state *j*. For a model λ, if O is an observation sequence $O = (O_1, O_2,...,O_T)$ which is assumed to have been generated by a state sequence *Q= (Q1, Q2,..,QT)*, of length *T*, we calculate the observations probability or likelihood as follows:

$$P(O, Q|\lambda) = \sum_Q \pi_{q1} b_{q1}(O_1) \prod_T a_{qT-1\ qT} b_{qT}(O_T) \quad \ldots\ldots\ldots (4)$$

Where $\pi_{q1}$ is initial probability of state 1. The classification is performed using the Viterbi algorithm. In Fig.5 we show some examples where text line images are identified using HMM-based classification system.

| English | Devanagari | Bangla |
|---|---|---|
| CSP curriculum and professional | परिंदों को उडान मुबारक | রাষ্ট্র ভাষা বাংলা চাই |
| The whole environment is quiet | करना उनकी मजबूरी हो जाये | বিনা প্রয়োজনে ঘুরাফেরা করবেন না |
| GSG has transformed how enterprises | स्वरूपानंद: शनि भगवान नहीं | রক্তদান জীবন দান |

**Fig.5. Some examples of English, Devanagari and Bangla text lines separated by our system**

### 3.2.2. Keyword-Translation

There exist a number of translators that translates phrases of words from one language to another. This is due to the increased levels of global communication in different geographical regions of the world. In our word-spotting framework, given a user keyword which was used to search documents from different scripts, we identify the script of the text line image and next the keyword is translated to that corresponding script. We have used Google Translate Services to translate the words. The choice of using Google translate is due to easily available to users and is free-of-cost with API available to users.

**Table I: Some English keywords and their translated words in Bangla and Devanagari used in word spotting.**

| English | Bangla | Devanagari | English | Bangla | Devanagari |
|---|---|---|---|---|---|
| College | কলেজ | ☐☐☐☐☐ | Water | ☐☐ | ☐☐☐☐ |
| School | ☐☐☐☐☐☐☐☐ | ☐☐☐☐☐☐☐☐ | Geography | ☐☐☐☐☐ | ☐☐☐☐☐ |
| Father | ☐☐☐☐ | ☐☐☐☐ | Research | ☐☐☐☐☐☐ | ☐☐☐☐☐☐☐☐ |
| Professor | ☐☐☐☐☐☐☐ | ☐☐☐☐☐☐☐☐☐☐ | Morning | ☐☐☐☐ | ☐☐☐☐ |
| History | ☐☐☐☐☐☐ | ☐☐☐☐☐☐ | Food | ☐☐☐☐☐ | ☐☐☐☐ |

### 3.3. Two-Stage Word Spotting Approach

In this section, we present our word 2-stage word spotting framework from different scripts. We first discuss the word-spotting framework and its details. Next, the two stage approaches are discussed. In the first stage, character classes are reduced using dynamic shape coding. Due to complex Indic character shapes. zone segmentation is performed and middle zone characters are grouped according to similarity. Next, in the second stage the word hypotheses are verified. These are detailed in following sub-sections. The flowchart our HMM-based word spotting framework is shown in Fig.6.

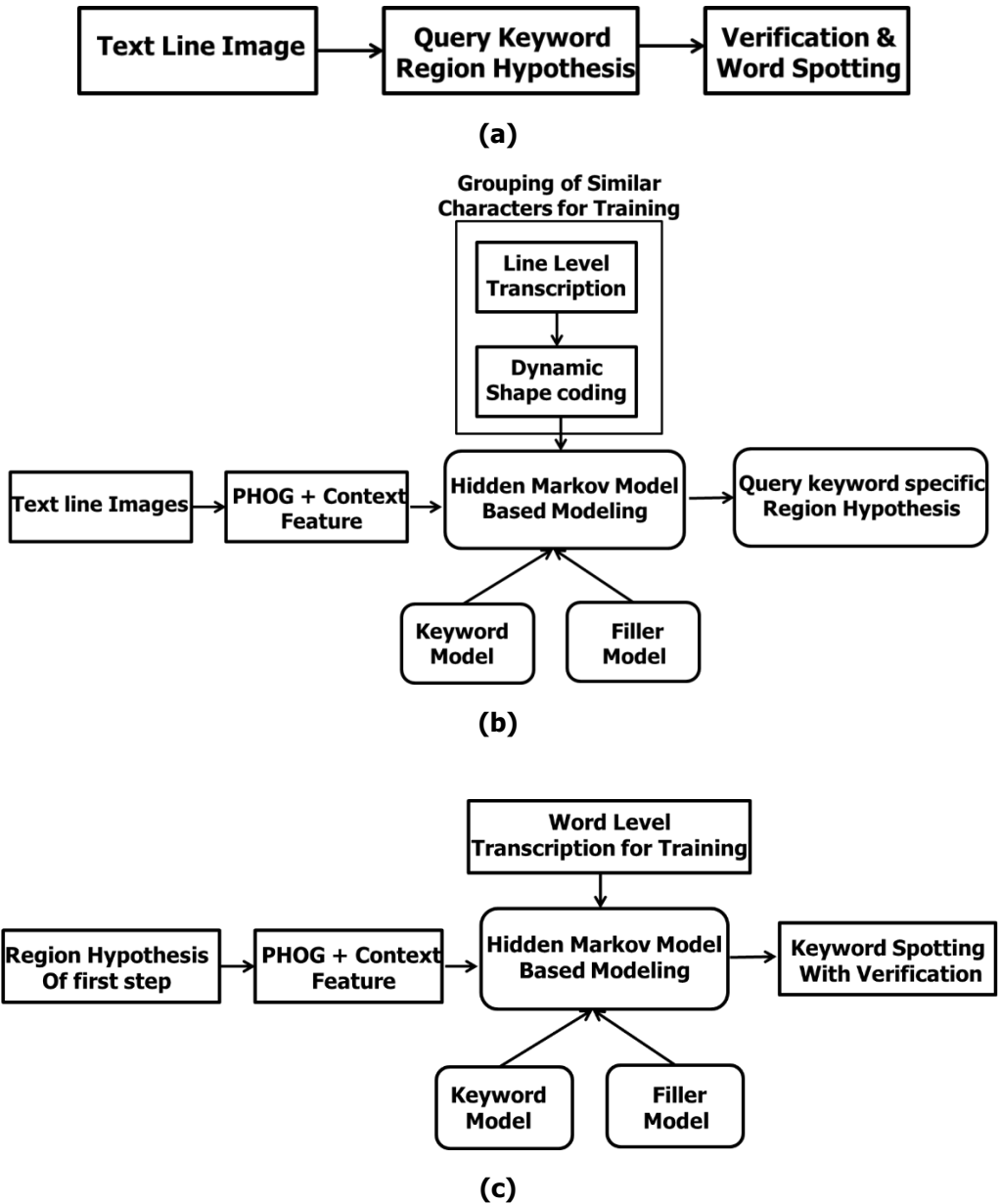

**Fig.6. Details of our word spotting framework. (a) Two stage framework for hypothesis location detection and verification. (b) In first stage, dynamic shape coding is used to reduce the character set and find the hypothesis detection. (c) Second stage uses the actual transcription to verify the regions.**

### 3.3.1. HMM-based Word Spotting Framework

In word spotting framework, during training, sliding window features are extracted from labelled text line images. The probability of the character model of the text line is then maximized by Baum-Welch algorithm assuming an initial output and transitional probabilities. Using the character HMM models, a filler model has been created [10].The keyword model which has been used in our system is to spot a keyword in a text line image. The filler model represents a single character model consisting of any one of the characters. A 'Space' model has been used in the keyword model for modelling white spaces.

Word spotting mechanism is based on the scoring of text image (X) for the keyword (W). If the score value is greater than a certain threshold then it gives a positive value for the occurrence of that particular keyword in that text line. The score assigned to the text line image X for the keyword W is based on the posterior probability P(W|$X_{a,b}$) trained on keyword models. Where a and b correspond to starting and ending position of the keyword whereas $X_{a,b}$ gives the particular part of text line containing the keyword [10]. Applying Bayes' rule we get

$$\log p(W|X_{a,b}) = \log p(X_{a,b}|W) + \log p(W) - \log p(X_{a,b}) \dots \dots \dots (5)$$

Considering equal probability we can ignore the term $\log p(W)$. The term $\log p(X_{a,b}|W)$ represents the keyword text line model where it is assumed that exact character sequence of the keyword to be present separated by 'Space'. The rest part of the text line is modelled with Filler text line model. Then we can find the position a, b for the keyword alongside with the log-likelihood $\log p(X_{a,b}|W) = \log p(X_{a,b}|K)$. $\log p(X_{a,b})$ is the unconstrained filler model F. The general conformance of the text image to the trained character models is given by obtained log-likelihood $\log p(X_{a,b}) = \log p(X_{a,b}|F)$. The difference between the log-likelihood value of keyword model and filler model is normalized with respect to the length of the word to get the final text line score.

$$Score(X, W) = \frac{[\log p(X_{a,b}|K) - \log p(X_{a,b}|F)]}{b-a} \dots \dots \dots (6)$$

Then this $Score(X, W)$ is compared with a certain threshold[10] for word spotting. To improve the word spotting performance we have included context feature from neighbour windows. It is briefly discussed as follows.

**Dynamic Features:** For developing an efficient word spotting system, an improved feature extraction in sliding window has been used in our work. This involves including contextual information from the neighboring windows by adding time derivatives in every feature vector. The combination of contextual and dynamic information [51] in the current window helps to improve the performance of a word spotting system. The first order dynamic features are known as delta coefficients while the second order dynamic feature are known as acceleration coefficients. The delta coefficients may be expressed in terms of first order regression of feature vector. The value of the delta coefficient may be calculated as:

$$c_t = \frac{\sum_{\emptyset=1}^{\theta} \emptyset(c_{t+\emptyset} - c_{t-\emptyset})}{2\sum_{\emptyset=1}^{\theta} \emptyset^2} \ldots \ldots \ldots (7)$$

In the equation written above, $c_t$ is a delta coefficient which has been computed in terms of the corresponding static coefficients $c_{t+\emptyset}$ and $c_{t-\emptyset}$. The value of $\theta$ is determined according to the size of the window. On the other hand, the value of the acceleration coefficients can be computed using second order regression. These features play an important role of capturing temporal information at each frame level and representing the dynamics of features around the current window. In this work, we have used 168 dimensional feature vector along with the delta and acceleration coefficients.

### 3.3.2. Stage-I: Word Spotting with Dynamic Shape Coding

Here, the word spotting approach using dynamic shape coding scheme has been discussed. In each script, usually, there are many characters which look similar in shapes, e.g. 'E' and 'F' in Latin, ' □ ' and '□ ' in Bengali, ' □ ', '□ ' in Devanagari. Due to these similar looking characters word spotting may fail to detect the query word properly because of confusion. Traditional word spotting approaches provide high score in similar looking characters which reduces the retrieval performance. To overcome this confusion, in the first stage, similar shape characters are grouped together in HMM-filler model. Grouping similar characters reduces the number of character classes in a script, thus, improves the detection of keywords. As a side product, the alignment of query word detection is improved. Due to this nature of grouping characters, the modified HMM keyword model will detect words which are not same, but the characters of the words will be of similar appearance. "EAT", "FAT" will be detected with the same query word due to similar character shape of 'E' and 'F'. This problem will be taken care during verification (second stage).

Incorporating shape coding based word spotting scheme provides advantages over traditional methods. Shape coding based text encoding approach has been used efficiently in documents [52], [53]. These approaches annotate character images by a set of predefined codes. Nakayama [53] annotated character images by seven shape codes for word content detection. Lu et al. [52]

proposed a set of topological codes based on shape features including character ascenders/descenders, holes, water reservoir information, etc. to retrieve document images. Usually, these shape codes were manually designed to perform text document searching which is script dependent and very tedious. In this paper, we propose grouping character shapes using a dynamic way which is unsupervised and can be extended to any script. The unsupervised grouping of character shapes provides flexibility in encoding characters. This is detailed in experiment section 4.3.3. The number of reduced character sets is adjusted according to validation data of the script.

Inspired with this idea, the proposed keyword spotting approach uses HMM-based word spotting using shape coding. Similar-shaped text characters are grouped together and they are encoded according to [52]. The word spotting approach is next modelled using HMM by modified shape coding based character models. Finally, words are searched by this shape coded character HMM approach.

We propose here a novel unsupervised character shape clustering for word spotting purpose. The objective of the character clustering is to find similar shapes of the characters of a script in an unsupervised way. For this purpose, we have used an efficient shape descriptor using Zernike moments [54] and next clustering of the characters in hierarchical way. These are detailed in following subsections. A flowchart describing the process of generating the dynamic character codebook is shown in Fig.7.

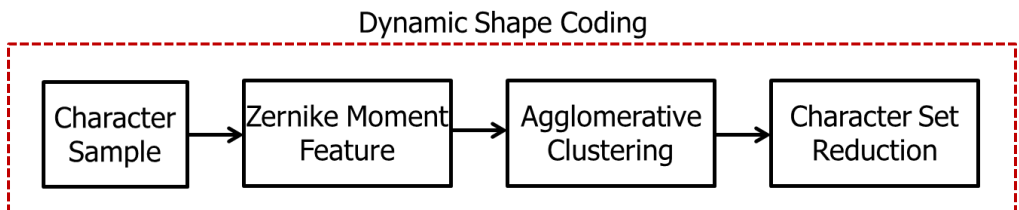

**Fig.7. The modules of generating dynamic character codebook.**

**Feature Extraction and Clustering:** Zernike moments [54] have been found to have minimal redundancy, rotation invariance and robustness to noise among moment based descriptors; therefore Zernike moments are used several applications on image analysis, reconstruction and recognition. Zernike moments are based on a set of complex polynomials that form a complete orthogonal set over the interior of the unit circle [54]. They are defined to be the projection of the image function on these orthogonal basis functions. The basis functions $V_{n,m}(x,y)$ are given by:

$$V_{nm}(x, y) = V_{nm}(r, \theta) = R_{nm}(\rho)e^{jm\theta} \quad \ldots\ldots\ldots (8)$$

where n is a non-negative integer, m is a non-zero integer subject to the constraints n-|m| is even and $n < |m|$, $\rho$ is the length of the vector from origin to $(x, y)$, $\theta$ is the angle between vector $\rho$ and the x-axis in a counter clockwise direction and $R_{n,m}(\rho)$ is the Zernike radial polynomial. The Zernike radial polynomials, $R_{n,m}(\rho)$, are defined as:

$$R_{nm}(\rho) = \sum_{s=0}^{(n-|m|)/2} \frac{(-1)^s (n-s)!}{s!\left[\frac{n+|m|}{2} - s\right]!\left[\frac{n-|m|}{2} - s\right]!} r^{n-2s} \quad \ldots\ldots\ldots (9)$$

Note that, $R_{n,m}(\rho) = R_{n,-m}(\rho)$. The basis functions in equation 1 are orthogonal thus satisfy,

$$\frac{n+1}{\pi} \iint_{x^2+y^2 \leq 1} V_{nm}(x,y) V_{pq}^*(x,y) = \delta_{np}\delta_{mq} \quad \text{where, } \delta_{ab} = \begin{cases} 1 & \text{if } a = b \\ 0 & \text{otherwise} \end{cases} \quad \ldots\ldots\ldots (10)$$

The Zernike moment of order n with repetition m for a digital image function $f(x,y)$ is given by

$$Z_{nm} = \frac{n+1}{\pi} \sum \sum_{x^2+y^2 \leq 1} f(x,y) V_{pq}^*(x,y) \quad \ldots\ldots\ldots (11)$$

where, $V_{nm}^*(x, y)$ is the complex conjugate of $V_{nm}(x, y)$.

To compute the Zernike moments of a given character image, the image centre of mass is taken to be the origin. In our approach, the character images are normalized into 41*41 before applying Zernike feature computation. The size is considered from the performance of experimental data.

**Agglomerative Hierarchical Clustering:** The Zernike moment feature calculated from character components from a set of training image are used for clustering. In order to handle clusters that can take any shape, we adopt the popular Single Linkage Agglomerative Clustering and apply it to the character features. This algorithm merges the two closest clusters into higher-level clusters, where the cluster distance is defined as the minimum inter-clusters character distance. The process stops when the distance between the closest clusters exceeds a predefined threshold or when all the characters have been merged into a single cluster. Some examples of similar character shapes for three scripts are shown in Table II. The output of the stage-I of our word spotting framework is shown in Fig.8.

Table II: Shape coding representation for different scripts

| | English | | | | Bangla | | | | Devanagari | | |
|---|---|---|---|---|---|---|---|---|---|---|---|
| **1.** | A | **20.** | b | **1.** | ☐ | **20.** | ☐, ☐ | **1.** | ☐ | **19.** | ☐, ☐ |
| **2.** | B, R, 8 | **21.** | c, o, e | **2.** | ☐, ☐ | **21.** | ☐ | **2.** | ☐, ☐ | **20.** | ☐, ☐ |
| **3.** | C, G | **22.** | f | **3.** | অ | **22.** | ☐ | **3.** | ☐, ☐ | **21.** | ☐, ☐ |
| **4.** | D, O, Q, 0 | **23.** | g | **4.** | ☐, ড, ☐ | **23.** | ☐, ☐ | **4.** | ☐ | **22.** | ☐, ☐ |
| **5.** | E, F | **24.** | i, j | **5.** | ☐, ☐ | **24.** | ☐ | **5.** | ☐, ☐ | **23.** | ☐, ☐ |
| **6.** | H | **25.** | m | **6.** | ☐ | **25.** | ☐ | **6.** | ☐ | **24.** | ☐ |
| **7.** | I, T, l, t, 1 | **26.** | n | **7.** | ☐, ☐ | **26.** | ☐ | **7.** | ☐ | **25.** | ☐ |
| **8.** | J, d | **27.** | q | **8.** | ☐, ☐ | **27.** | ☐ | **8.** | ☐, ☐ | **26.** | ☐, ☐ |
| **9.** | K, k | **28.** | r | **9.** | ☐ | **28.** | ☐, ☐ | **9.** | ☐ | **27.** | ☐ |
| **10.** | L | **29.** | u, v | **10.** | ☐ | **29.** | ে | **10.** | ☐ | **28.** | ☐, ☐ |
| **11.** | M | **30.** | y | **11.** | ☐ | **30.** | ☐, ☐ | **11.** | ☐, ☐ | **29.** | ☐ |
| **12.** | N | **31.** | X, x | **12.** | ☐ | **31.** | ঃ | **12.** | ☐ | **30.** | ☐ |
| **13.** | P, p, 9 | **32.** | 2 | **13.** | ☐, ☐ | **32.** | ং | **13.** | ☐, ☐ | **31.** | ☐ |
| **14.** | S, s | **33.** | 3 | **14.** | ☐ | **33.** | া | **14.** | ☐, | **32.** | ☐ |
| **15.** | U, V | **34.** | 4 | **15.** | ☐ | **34.** | ☐ | **15.** | ☐, ☐ | **33.** | ☐ |
| **16.** | W, w | **35.** | 5 | **16.** | ☐ | **35.** | ☐ | **16.** | ☐ | **34.** | ❖ |
| **17.** | Y | **36.** | 6 | **17.** | ও | **36.** | ☐ | **17.** | ☐, ☐ | **35.** | T |
| **18.** | Z, z, 7 | | | **18.** | ☐, ☐ | **37.** | ☐ | **18.** | ☐, ☐ | | |
| **19.** | a | | | **19.** | ☐, ☐, ☐ | **38.** | ☐ | | | | |

| Query Keyword | Text line Images | Result |
|---|---|---|
| Fat | Fat free oil from Amul!! Stay healthy | TP |
| | Eat healthy, think better. | FP |
| local | Radio mirchi means total entertainment | FP |
| | The best local made coffee. | TP |
| Coal | Coal is precious, do not waste. | TP |
| | Goal is set to be achieved | FP |

**Fig. 8:** Examples showing keyword region hypothesis obtained from stage-I of our word spotting framework for different given query keywords. Actual result of prediction is given in a separate column. TP and FP stand for true positive and false positive.

**Character Shape Reduction in Indic Script:** As discussed earlier, full-zone wise character based HMM models may not be found fruitful in Indian scripts, especially in Bangla and Devanagari. As mentioned earlier, characters of most of the Indic scripts are written in upper, middle and lower zones. With morphological combination of characters with modifiers, the number of character classes becomes huge. Hence, sufficient data for each class will be necessary for training the respective class models. To deal with this problem, zone segmentation based word recognition approach has been adapted in [3], [43]. This approach reduces the character classes drastically and makes it robust to model the character classes using lesser training data along with major improvement in recognition accuracy. In Fig.9, it has been shown how zone segmentation reduces the number of character class in Bangla and Devanagari scripts using an example.

**Fig.9:** Examples showing character (middle zone) unit reduction using zone segmentation for Devanagari character 'क' and Bangla character 'ক' which are combined with modifiers.

To perform the zone segmentation approach in Indic script, the first step is to detect the proper region of Matra. In printed Indic scripts, usually the row with highest peak in horizontal projection analysis detects the Matra. In literature, projection based analysis[5] has been used for segmenting upper and lower zones in printed text. We have used similar projection analysis approach for segmentation of three zones in Indic scripts. These segments which move upwards are considered as upper-zone components of the word image. Similarly, lower-zone components are separated from lower part of the projection analysis [55]. The middle portion of the text lines is considered as middle-zone components. Fig.10 shows examples of zone segmentation on Bangla text line. Since, the components in middle zone convey maximum information of the text, these components are used for word spotting. The verification is performed finally after considering topological information of upper and zone components of spotted word location.

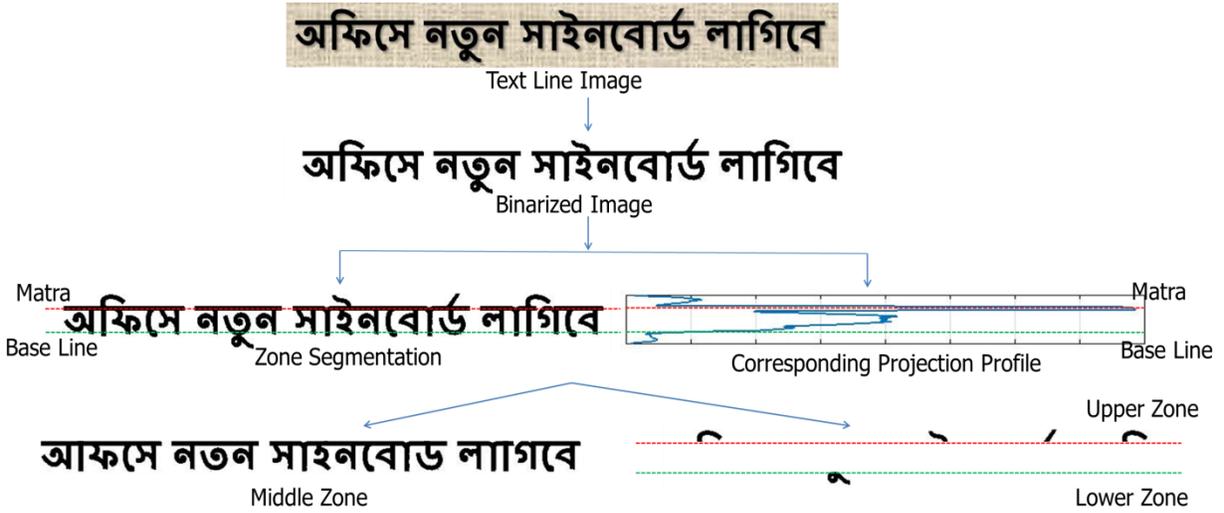

**Fig.10. Examples showing zone segmentation on Bangla text-line images**

### 3.3.3. Stage-II: Verification of Query Keywords

In this stage, the hypotheses locations are verified using word spotting approach. During the training, the original transcription of the characters in text lines is used for training. Dynamic context feature has been included for feature extraction. Next, the text location obtained from first stage is passed through word spotting approach for rejecting the false alarms. The filler models for the both the stages are shown in Fig. 12. Note that, the number of characters in first set is less which improves the Viterbi based forced alignment. Fig. 11 shows some examples of our verification stage. An algorithm of our proposed framework is also provided(see Algorithm-1).

| Query Keyword | Word region hypothesis obtained from Stage-I | Verification Output | Result |
|---|---|---|---|
| Fat | Fat | Positive | ✓ |
|  | Eat | Eliminated |  |
| local | total | Eliminated | ✓ |
|  | local | Positive |  |
| Coal | Coal | Positive | ✓ |
|  | Goal | Eliminated |  |

**Fig. 11:** Examples showing verification step of our word spotting framework corresponding to Fig. 8. False positive cases are eliminated using this verification stage.

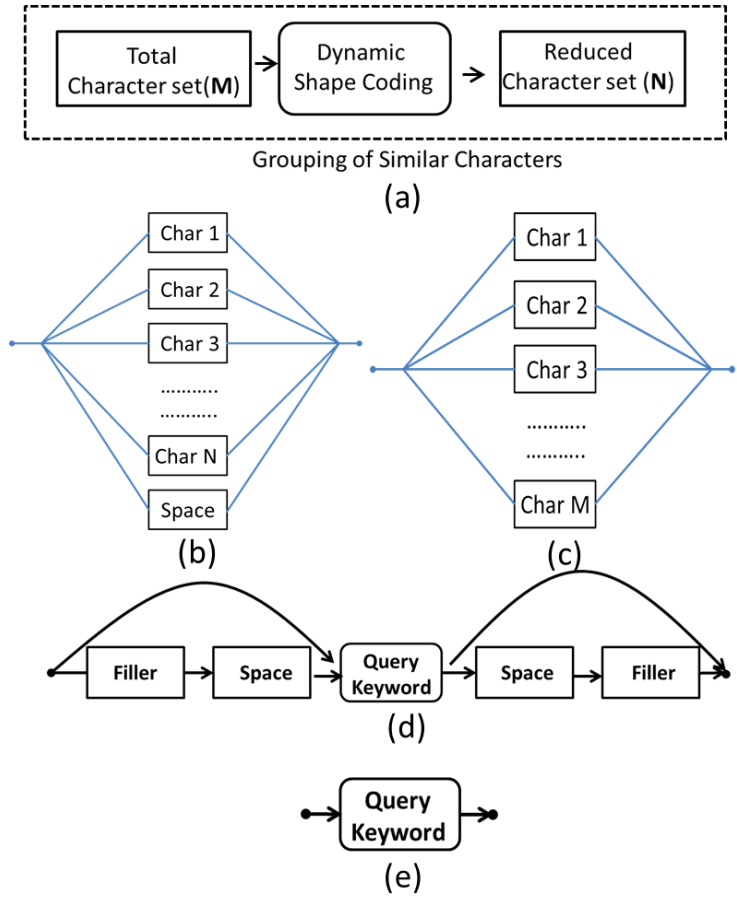

**Fig. 12:** (a) Total character set (M) is converted to reduced character set using dynamic shape coding. (b) Filler model for stage-I which uses reduced character set (N) along with 'Space' unit. (c) Filler model for verification stage which uses complete character set(M) with exclusion of Space unit as dealing with cropped query keyword region obtained from stage I. (d) Keyword model for stage -I (e) Keyword model for verification stage.

**Algorithm1.** Word Spotting in multiple scripts using Dynamic Shape Coding

**Input:** Text line images with a given query keyword $K_W$
**Output:** Word spotting in multiple scripts.

**for** each text-line $T_i$ of $T_1...T_N$ **do**

    Step 1: Binarize $T_i$ along with enhancement. (Section 3.1)

    Step 2: Identify the script ($L_i$) of the text line $T_i$. (Section 3.2)

    Step 3: Translate $K_W$ into language $L_i$

    Step 4: Encode $K_W$ into $K_{W|D}$ using Dynamic Shape Coding in order to group similar characters. (Section 3.3.2)

Step 5: Extract PHOG and Context feature from $T_i$ to accomplish the HMM-based modelling.

Step 6: Generate query keyword region hypothesis $(W_{a,b})$ for every text lime image$(T_i)$ using $K_{W|D}$ where a and b stand for starting and ending boundary of query keyword. (Section 3.3.2)

Step 7: Verify $(W_{a,b})$ using $K_W$ (Section 3.3.3)

**end for**

## 4. Experiment Results & Discussions

In this section, we present the performance of our script independent word searching framework. As there exists no such publicly available datasets for scene/video text images for Devanagari and Bangla script, we have collected our own dataset to measure the performance. There exist a number of publicly available datasets for English. We have collected both scene images and videos from those datasets along with our own additions for English. To justify the performance of our dynamic shape coding based word spotting framework, we have shown the advantages of the proposed two-stage word spotting approach in publicly available English datasets as well as our own private datasets.

### 4.1. Dataset

For our experiment, we have collected a total of 956 scene images and 48 video files from news-channels clippings collected from YouTube for Devanagari script. 3,341 scene text line images and 3,086 video text line images are collected from 956 scene images and 48 video files for Devanagari script respectively. Text line segmentation is done using [35] from scene images and video frames. For Bangla (English) script, a total number of 3,245(3,642) scene text line images and 2,898(3,485) video text line images are collected from 848(912) scene images and 46(52) video files respectively. Fig.13 shows the examples of different scene and video frames considered for our dataset. We divide our datasets into three parts as training, testing and validation. Different parameters of our experiment are evaluated using validation dataset. During the verification stage of word spotting framework, we train the HMM-model using word images along with word level transcription. For this, we have collected a total of 5K training word sample for English script from publicly available scene text recognition dataset. In case of Devanagari and Bangla scripts, we separately collected a total of 4,782 and 4,861 word images to train the HMM-model used in the verification stage. These word images of corresponding script are used to train the HMM-model of verification stage both in case of video and scene frames. The complete description of our dataset along with part-wise division for training, testing and validation is tabulated in Table III.

Table III: Description of data details for experiment evaluation

| Script | | Training | Testing | Validation |
|---|---|---|---|---|
| Devanagari | Scene Image | 1825 | 808 | 708 |
| | Video Frame | 1630 | 746 | 710 |
| Bangla | Scene Image | 1717 | 812 | 716 |
| | Video Frame | 1498 | 705 | 695 |
| English | Scene Image | 1875 | 911 | 856 |
| | Video Frame | 1789 | 894 | 802 |

| Type | English | Bangla | Devanagari |
|---|---|---|---|
| Scene Image | 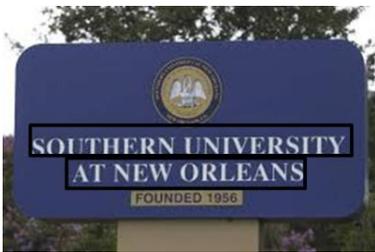 | 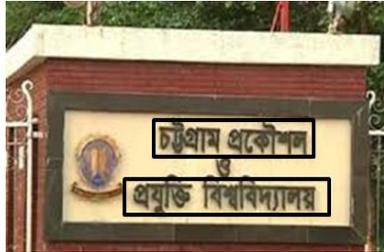 | 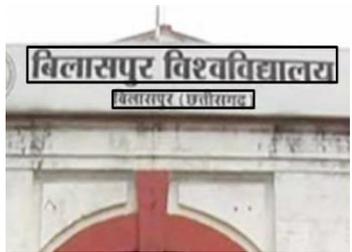 |
| Video Image | 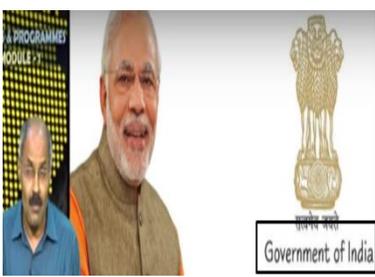 | 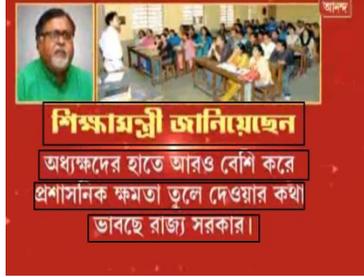 | 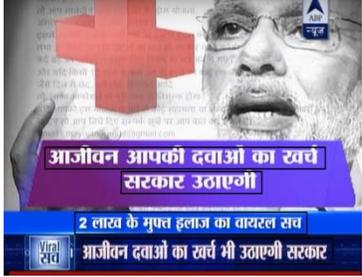 |

**Fig.13. Examples showing scene images and video frame of our datasets.**

We have measured the performance of our word spotting system using precision, recall and mean average precision (MAP). The precision and recall are defined as follows.

$$Precision = \frac{TP}{TP+FN} \quad Recall = \frac{TP}{TP+FP} \ldots\ldots\ldots (12)$$

Where, TP is true positive, FN is false negative and FP is false positive. F-measure is given by the harmonic mean of average precision and recall value.

## 4.2. Performance of Script Identification

For our experiment, text line images from scene image and video frames are considered as discussed in Section 4.1. After extracting the text lines, the script classification is performed using HMM. We have used PHOG feature for this purpose. We performed our script identification experiment on 10,334 samples (of three types) as training data, 4,876 samples as testing data and rest 4,487 for validation purpose. We achieved an identification accuracy of 98.94% using PHOG features. Fig.14 shows some qualitative results where our approach identifies the script of the lines correctly. We have compared our script identification framework for text line images with popular LGH (Rodriguez-Serrano & Perronnin, 2009) (Local Gradient Histogram) feature. Here, similar to PHOG, a sliding window is being shifted from left to right of the word image with an overlapping between two consecutive frames. The size of the feature dimension is 128. PHOG feature provides an improvement of 0.78% accuracy over LGH in script identification. A confusion matrix corresponding to the script identification is shown in the Fig.15.

| Text Line Image | Ground tooth | Result |
|---|---|---|
| 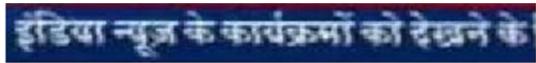 | Devanagari | ✓ |
| 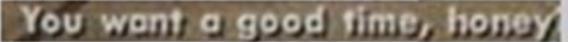 | English | ✓ |
| 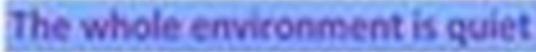 | English | ✓ |
| 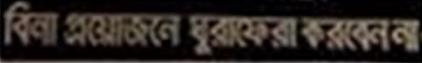 | Bangla | ✓ |
| 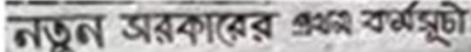 | Bangla | ✓ |
| 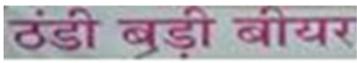 | Devanagari | ✓ |
| 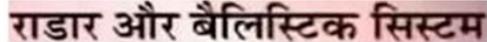 | Devanagari | ✓ |
| 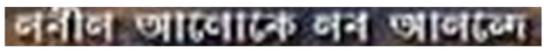 | Bangla | ✓ |

**Fig.14. Some examples of qualitative results where our system identified the text lines correctly.**

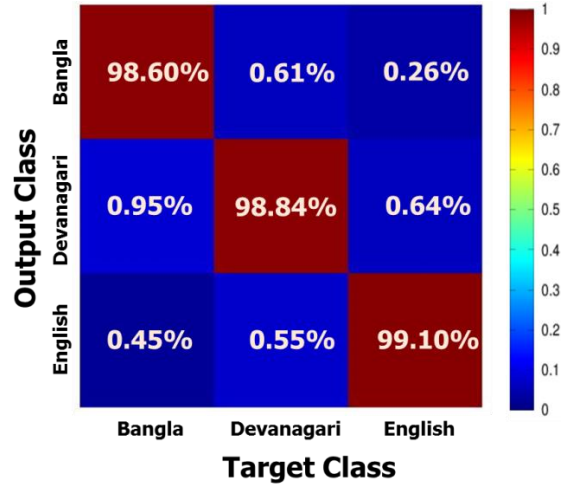

**Fig.15. Confusion matrix for line level classification.**

## 4.3. Key-Word Retrieval Performance

The training is performed with the popular HTK toolkit [57]. We used continuous density HMMs with diagonal covariance matrices. The performance of our proposed methodology is measured by including successive modules in our system.

### 4.3.1. Performance of our Two-stage Word Spotting framework

The color images of text lines are converted to Binary image using Bayesian classification as described in Section 3.1. Next, the binary text lines are searched for the query word after identifying the script. The translated query word is used to search the query keyword in the target text lines. Dynamic shape coding is used to group the similar characters in order to reduce the confusion during detection of the query keyword. We generate query keyword region hypothesis from the first stage of our word spotting framework. During verification stage, we train the HMM-model using word images and original word-level transcription, rather than training from text line images in order to make it more robust. Here, we have considered the method in [10] as the baseline method, named as Filler-HMM, to evaluate the improvement of our proposed word spotting framework. During training, we noticed that 32 Gaussian mixture models and 8 states in HMM set-up provided best results from validation dataset. Next, we evaluated results on test dataset. During testing, the spotting results are compared with the ground-truth data to compute the accuracy. The log-likelihood scores are thresholded by a global threshold value [10] to distinguish between correct and incorrect spotting. To get an idea of the word spotting results, some qualitative results, some of the keywords that have been spotted by our system along with incorrect spotting have been shown in Fig.16. Note that, the zone segmentation is not included in this experiment. Some of the spotting results are incorrect due to complex nature of Indic script.

| Query Keyword | Script | Translated Keyword | Text Line Image | Result |
|---|---|---|---|---|
| Professor | English | Professor | Associate professor is an academic title that | ✓ |
| | Bangla | অধ্যাপক | [Bangla text line] | ✗ |
| | Devanagari | प्रोफ़ेसर | पदभार के रूप से प्रोफ़ेसर नही होते | ✓ |
| History | English | History | History faculty win grants from the | ✓ |
| | Bangla | ইতিহাস | ১৯৪৭-পরবর্তী ভারতীয় প্রজাতন্ত্রের ইতিহাস জানতে | ✓ |
| | Devanagari | इतिहास | परिहास और उपहास में अंतर करने का | ✗ |
| Museum | English | Museum | Welcome to the MIT Museum | ✓ |
| | Bangla | জাদুঘর | [Bangla text line] | ✗ |
| | Devanagari | संग्रहालय | भारतीय संग्रहालय पश्चिम बंगाल के | ✓ |

**Fig.16. Qualitative results with correct as well as incorrect spotting.**

To justify our two stage word spotting framework using dynamic shape coding, we have evaluated the performance of the first stage our framework where our intension is to find the course level word retrieval details with an emphasize on finding the location of the query keyword in a test line image. As stated in section 3.3.2, grouping of similar characters may sometimes result in false positive cases due to similar appearance of the grouped characters. During verification stage, these false positive results are eliminated using HMM-model trained from word images with original word level transcription. This successive improvement is shown through a precision-recall curve in Fig 17. This also includes the performance of traditional HMM-filler model based word spotting in our different datasets where query keyword is spotted in a single stage without including any shape coding method. Please note that we have not used zone segmentation for word spotting in Indic scripts in Fig. 17. The improvement obtained using zone segmentation in case of Indic scripts is provided in section 4.3.2. In this present section 4.3.1, we used our two stage word spotting framework along with dynamic shape coding method to evaluate the precision-recall values. We have also obtained a little improvement in the average precision and recall values since we are using contextual information from the neighboring windows to train the HMM model for word spotting. The improvement using context features is tabulated in Table IV.

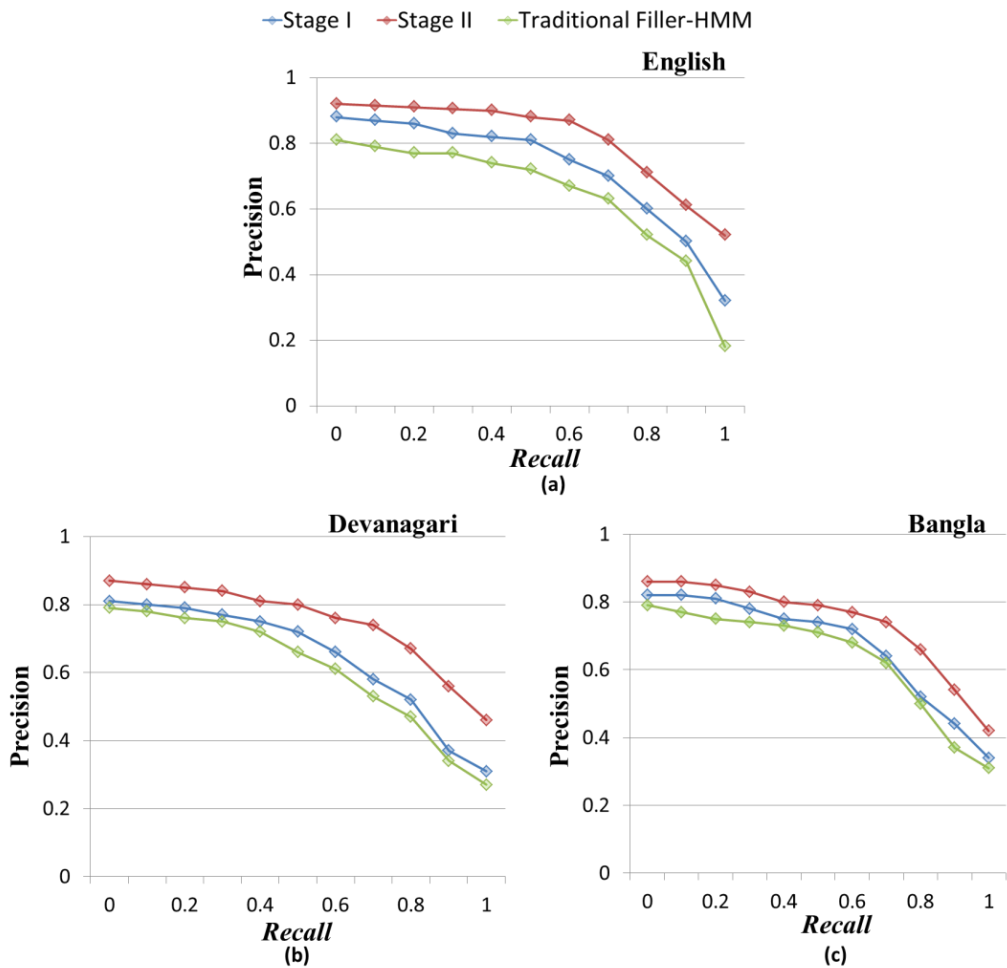

Fig.17. Precision-Recall curve using showing the improvement over traditional filler-HMM model in different scripts, namely, English, Devanagari, Bangla.

Table IV: Comparative study of F-measure with and without context feature

| Script | Average F-Measures | |
|---|---|---|
| | Without context features | With context feature |
| English | 65.32 | 67.00 |
| Bangla | 55.94 | 57.99 |
| Devanagari | 56.12 | 58.25 |

### 4.3.2. Improvement in Indic script using zone segmentation

We have next analysed the performance of our system including zone segmentation module as discussed in Section 3.3.2. Due to the complex shapes of Indic characters spread over three zones the traditional word spotting approach does not perform well. After segmenting the zones of Indic text, the word spotting performance improves. To evaluate the word-spotting performance in Indic scripts using zone segmentation, we have compared with the same without zone-segmentation analysis. Both these experiments, without and with zone segmentations, are performed using HMM along with dynamic shape coding method. The comparative studies in Bangla and Devanagari scripts are shown in Fig.18. Note that middle-zone based spotting performance shows improvement in both Bangla and Devanagari scripts with a high margin. The precision-recall and F-measure values are shown in Table V.

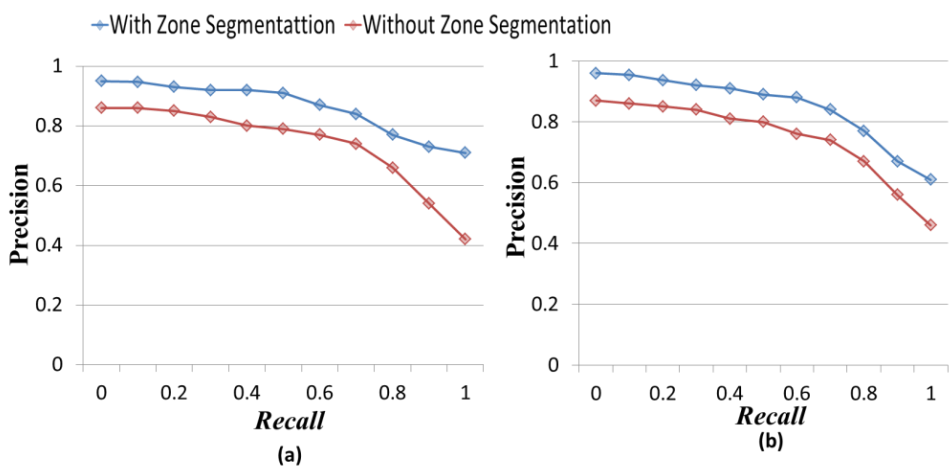

**Fig.18.** Precision-Recall curve using with and with-out zone segmentation based approach in (a) Bangla and (b) Devanagari scripts.

Table V: Average Precision-Recall values for both with and without zone segmentation based approach.

| Method | Script | Precision | Recall | F – Measure |
|---|---|---|---|---|
| Without Zone Segmentation | Bangla | 57.36 | 58.64 | 57.99 |
| | Devanagari | 57.69 | 58.84 | 58.25 |
| With Zone Segmentation | Bangla | 73.36 | 74.21 | 73.78 |
| | Devanagari | 73.69 | 73.92 | 73.80 |

### 4.3.3. Significance of dynamic shape coding

Next, we have integrated shape coding based scheme in word spotting. Similar character shapes are combined using Single Linkage Agglomerative Clustering to reduce the character confusion and hence word detection is improved. It is due to merging of similar shaped characters together. In Fig.19, we show dendograms of three different scripts. The lines marked in red indicate the sampling of characters at different levels. Table VI shows the grouping of characters combined in three different levels ($X_{+1}$, $X_0$, $X_{-1}$) indicated in Fig.19. It should be noted that the higher the level of the character combination (level $X_{+1}$) considered, the lesser will be the shape coding. In Table VII, the performance measures of corresponding shape codes at each level are detailed. The best result has been found by performing word spotting using level $X_0$ which is obtained from our validation dataset. The *F measures* and corresponding precision and recall values are shown in Table VIII for both with and without shape coding based method. A comparative study of qualitative results with and without shape coding based word spotting is shown in Fig. 20. With the help of shape coding, the confusion reduced and we obtained better performance in noisy images.

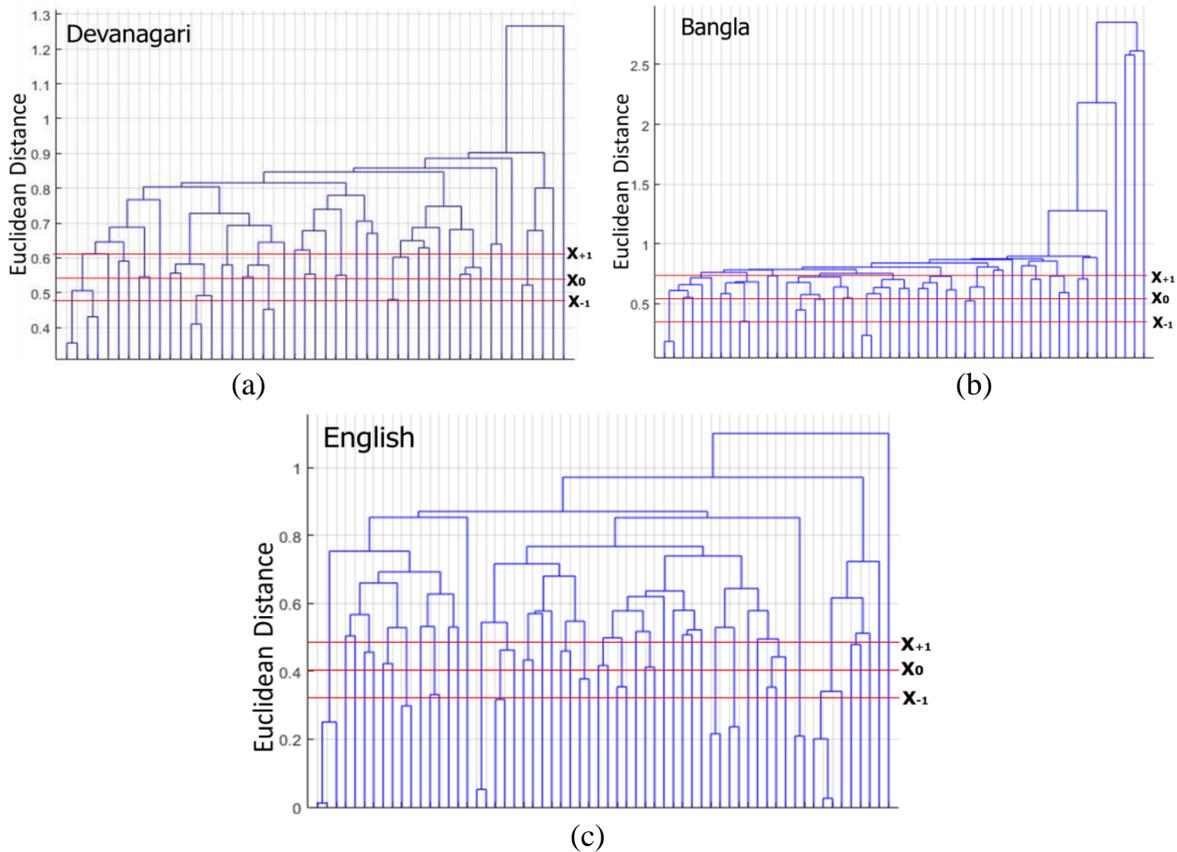

**Fig.19. Different dendogram for three scripts are shown. (a) Devanagari (b) Bangla and (c) English. Red lines indicate the sampling of characters at different level.**

**Table VI: Shape coding representation at different distance level for English**

| | | | | | | | | | | | | |
|---|---|---|---|---|---|---|---|---|---|---|---|---|
| colspan="13" | $X_{-1}$ |||||||||||
| 1. | A | 9. | F | 17. | N | 25. | Z | 33. | i | 41. | X, x |
| 2. | B, R | 10. | I, 1, t, l | 18. | P | 26. | z | 34. | j | 42. | 2 |
| 3. | 8 | 11. | T | 19. | p | 27. | 7 | 35. | m | 43. | 3 |
| 4. | C, G | 12. | J | 20. | 9 | 28. | A | 36. | n | 44. | 4 |
| 5. | D | 13. | D | 21. | S, s | 29. | B | 37. | q | 45. | 5 |
| 6. | O, 0 | 14. | K, k | 22. | U, V | 30. | c, e | 38. | r | 46. | 6 |
| 7. | Q | 15. | L | 23. | W, w | 31. | O | 39. | u, v | 47. | 9 |
| 8. | E | 16. | M | 24. | Y | 32. | G | 40. | y | 48. | |

| | | | | | | | | | | | | |
|---|---|---|---|---|---|---|---|---|---|---|---|---|
| colspan="13" | $X_0$ |||||||||||
| 1. | A | 7. | I, T, l, t, 1 | 13. | P, p, 9 | 19. | a | 25. | m | 31. | X, x |
| 2. | B, R, 8 | 8. | J, d | 14. | S, s | 20. | b | 26. | n | 32. | 2 |
| 3. | C, G | 9. | K, k | 15. | U, V | 21. | c, o, e | 27. | q | 33. | 3 |
| 4. | D, O, Q, 0 | 10. | L | 16. | W, w | 22. | f | 28. | r | 34. | 4 |
| 5. | E, F | 11. | M | 17. | Y | 23. | g | 29. | u, v | 35. | 5 |
| 6. | H | 12. | N | 18. | Z, z, 7 | 24. | i, j | 30. | y | 36. | 6 |

| | | | | | | | | | | | | |
|---|---|---|---|---|---|---|---|---|---|---|---|---|
| colspan="13" | $X_{+1}$ |||||||||||
| 1. | A | 6. | H | 11. | P, p, 9, 4 | 16. | Z, z, 7 | 21. | i, j | 26. | X, x |
| 2. | B, R, 8, 2 | 7. | I, T, l, t, 1, L, f | 12. | S, s | 17. | a | 22. | m, n, r | 27. | 3 |
| 3. | C, G | 8. | J, d | 13. | U, V | 18. | b, 6 | 23. | q | 28. | 5 |
| 4. | D, O, Q, 0 | 9. | K, k | 14. | W, w | 19. | c, o, e | 24. | u, v | 29. | |
| 5. | E, F | 10. | M, N | 15. | Y | 20. | g | 25. | y | 30. | |

**Table VII: Comparative results with different Euclidean distance during dynamic shape coding**

| Method | Script | Precision | Recall | F – Measure |
|---|---|---|---|---|
| English | $X_{+1}$ | 71.02 | 70.69 | 70.85 |
| | $X_0$ | 74.23 | 75.01 | 74.61 |
| | $X_{-1}$ | 69.98 | 68.89 | 69.43 |
| Bangla | $X_{+1}$ | 71.36 | 70.12 | 70.73 |
| | $X_0$ | 73.36 | 74.21 | 73.78 |
| | $X_{-1}$ | 68.94 | 68.21 | 68.57 |
| Devanagari | $X_{+1}$ | 71.32 | 71.12 | 71.21 |
| | $X_0$ | 73.69 | 73.92 | 73.80 |
| | $X_{-1}$ | 68.35 | 68.42 | 68.38 |

**Table VIII: Comparative Precision, Recall and F-measure with and without shape coding based method**

| Method | Script | Precision | Recall | F – Measure |
|---|---|---|---|---|
| Without shape coding | English | 68.85 | 67.19 | 67.00 |
| | Bangla | 68.54 | 67.14 | 67.83 |
| | Devanagari | 68.11 | 67.84 | 67.97 |
| With shape coding | English | 74.23 | 75.01 | 74.61 |
| | Bangla | 73.36 | 74.21 | 73.78 |
| | Devanagari | 73.69 | 73.92 | 73.80 |

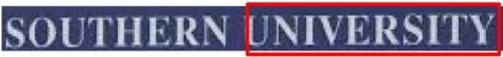

**Fig.20:** Example showing comparative study of word spotting performance using (a) with-out dynamic shape coding and (b) with dynamic shape coding

Using dynamic shape coding, we achieved a better query keyword boundary location compared to traditional filler-HMM [10]. To our knowledge, existing work did not evaluate the performance of word spotting in terms of boundary region of the query keyword. To measure the alignment, we define an evaluation metric for finding correctness in locating the boundary region for word spotting in text line images. Let, $L_{GT}^{start}$ and $L_{GT}^{end}$ be the actual starting and ending location for any query keyword. These ground truth locations, both $L_{GT}^{start}$ and $L_{GT}^{end}$, were marked manually for the query keywords in our datasets. Let, $L_{pred}^{start}$ and $L_{Pred}^{end}$ are the predicted starting and ending location of the query keyword. The percentage of error can be calculated by the eq. (13). This error value increases whenever the predicted boundary location includes unnecessary extra region other than the keyword location or it constricts than the usual ground truth location. The performance results based on error for with and without dynamic shape coding based approaches are given in Table IX. Fig. 21 describes the method used to evaluate the percentage of error for query keyword boundary region estimation. Some qualitative results for query keyword boundary estimation are shown in Fig. 22.

$$\% \, Error = \frac{|L_{GT}^{start} - L_{pred}^{start}| + |L_{GT}^{end} - L_{Pred}^{end}|}{|min(L_{GT}^{start}, L_{pred}^{start}) - \max(L_{GT}^{end}, L_{Pred}^{end})|} \ldots \ldots \ldots (13)$$

**Table IX: Percentage error in predicting the query keyword boundary location**

| Script | % Error (from eqn. - 13) | |
|---|---|---|
| | Without Dynamic Shape Coding | With Dynamic Shape Coding |
| English | 27.36 | 10.34 |
| Bangla | 32.12 | 14.36 |
| Devanagari | 31.14 | 13.69 |

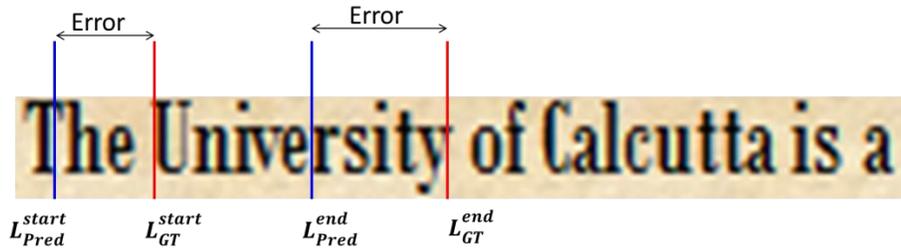

**Fig. 21 Example showing the error obtained in keyword boundary region detection, where red lines are the actual ground truth location and blue lines are predicted keyword boundary.**

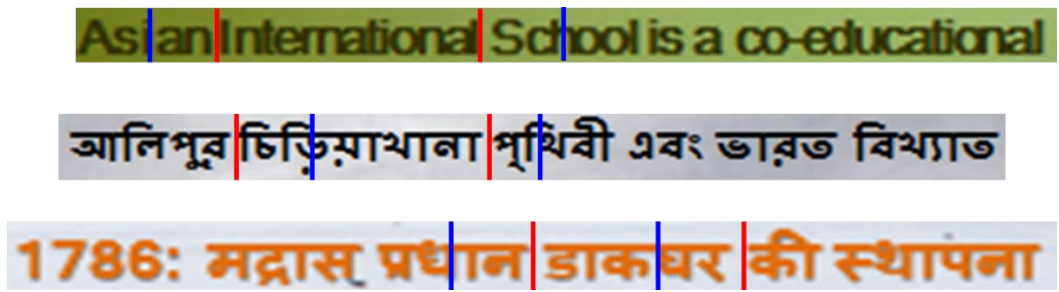

**Fig. 22 Example showing the boundary estimation obtained with and without using dynamic shape coding. Red lines denote the boundary estimation using dynamic shape coding and blue lines denote the boundary region obtained without using dynamic shape coding. Query keywords for these three consecutive text line images are 'International' (English), '☐☐☐☐☐☐☐☐'(Bangla) and '☐☐☐☐'(Devanagari) respectively.**

## 4.4. Parameter Evaluation

Continuous density HMM with diagonal covariance matrices of GMMs in each state has been considered for our framework. The window slides with 50% overlapping in each position. We have evaluated the performance of framework by varying the parameters. Experiments are carried out by varying the number of states from 4 to 8. A number of Gaussian mixtures were tested on validation data. The numbers of Gaussian distributions were considered from 16 to 128 increasing with a step of power of 2. Word searching performance is found to be optimum for all features with 32 Gaussian mixture and 6 as state number. Word spotting performance with different Gaussian numbers is detailed in Fig.23(a). Fig.23(b) illustrates the performance with varying the state number on word spotting experiment.

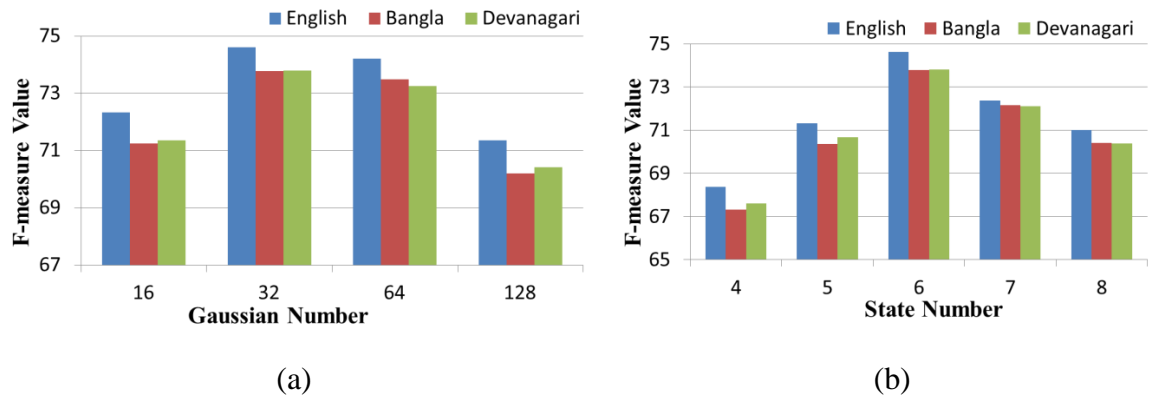

(a)　　　　　　　　　　　　　　　　(b)

**Fig.23: Word spotting performance evaluation using different (a) Gaussian number and (b) State number**.

## 4.5 Comparative Study

To compare the proposed word retrieval framework, we have tested with existing traditional features like Marti & Bunke [50], and LGH feature (Rodriguez-Serrano & Perronnin, 2009). The feature due to Marti & Bunke considered from foreground pixels in each image column. The fraction of foreground pixel information, the centre of gravity, moment feature and local features comprise of profile information were considered. The feature due to LGH (Rodriguez-Serrano & Perronnin, 2009), was similar to HOG feature [58] for object recognition. From each sliding window, a feature vector considering local gradient information was considered. Note that the proposed framework outperforms the other two features. From this result, we can infer that the performance of Marti & Bunke [50] feature is poor in our scene/video dataset since it captures only profile information for character modelling. We have also compared our method with recently proposed deep learning based architecture named PHOCNet [59]. However, please note that PHOCNet pipeline works on word level image, but our framework deals with the line level text line image which does not required any word level segmentation. Hence, we have used PHOCNet in place of our HMM-filler model based verification stage (see section 3.3.3) where the estimated keyword region proposals from the stage I are given as the input. From our experiments, we have found that HMM based verification stage exceeds the performance of PHOCNet in our all the datasets. Although deep learning based architecture achieved superior performance in many vision related tasks, the major limitation of this deep learning based paradigm is the requirement of a large dataset to train the model. Since there exists no large dataset for Indic scripts, the performance of deep network is limited in our case. To match the parity, we have only considered our own dataset for Latin script (English) even if some large datasets are available for it, and thereby limited performance is observed in English as well.

A comparative study with popular "Tesseract" OCR[1] is performed in case of English script. Here, English scene/video text line images are fed into OCR which extracts and recognize text. It performed well for images with high resolution. However, it often failed for images with low resolution, blur texts, complex background, etc. Table X shows the comparative study of our proposed method with Marti-Bunke feature, LGH feature, PHOCNet[59] and "Tesseract"-OCR.

**Table X: Comparison of word retrieval performance (using F-measure) with different features**

| Script | Marti-Bunke | LGH | Tesseract-OCR | PHOCNet [59] | Proposed |
|---|---|---|---|---|---|
| English | 61.58 | 71.64 | 51.36 | 71.69 | 74.61 |
| Bangla | 57.62 | 70.39 | - | 70.37 | 73.78 |
| Devanagari | 57.36 | 70.52 | - | 70.54 | 73.80 |

## 4.6 Error Analysis

From experiment, we found few images which weren't recognized properly by our binarization method due to low resolution, blur, and complex color transition. Some examples of improper binarization are shown in Fig.24. In videos and scene images, sometimes we encounter different fonts than the conventional ones. Because of cursive/stylish fonts, it may be difficult to spot any text of such font in video frames. For getting precise spotting even with various fonts, single character of different fonts should be embedded inside the character HMMs by training all possible fonts which seems to be tedious as there is a large number of fonts used by television channels, video editors, etc.

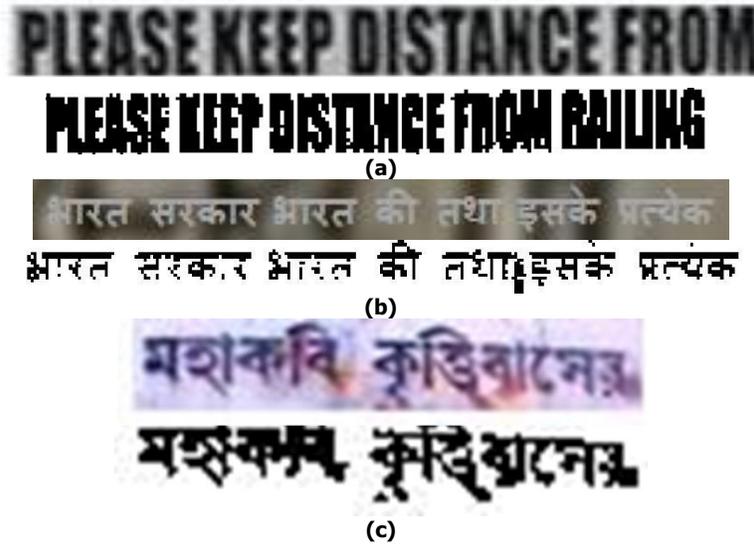

(a)

(b)

(c)

**Fig.24: Poorly binarized image where our word spotting system failed.**

## 4.7. Time Computation

Experiments have been done on an I5 CPU of 2.80 GHz and 4G RAM 64-bit Computer. For each query, the average runtime has been computed from different runs made in the experiment. In Table XI, we show the average time taken using queries of various approaches developed in MATLAB. Here, filler-HMM stands for the single stage HMM-based word spotting approach as proposed in [10]. We have used PHOG feature for all the experiments mentioned in Table XI. From Table XI, we can easily infer the effect of additional introduction of dynamic/context features and dynamic shape coding in our framework. Dynamic features increases the runtime whereas, runtime performance is improved using dynamic shape coding. Total runtime of our proposed two-stage word retrieval system is 1.82 seconds. Due to use of verification stage, overall run time of our proposed method is higher by 0.77 second compared to filler-HMM approach. However, we assume that the improvement in terms of average F-measure value and keyword boundary region at the cost of this slightly higher runtime is justified. To improve the time performance further, parallel processing or multicore methods [66] could be adapted.

**Table XI: Time computation analysis of date spotting using different approaches.**

|  | Method | Time (In Sec.) |
|---|---|---|
| Stage- I | Filler-HMM | 1.05 |
|  | Filler-HMM + Dynamic Feature | 1.19 |
|  | Filler-HMM + Dynamic Shape Coding | 0.71 |
|  | Filler-HMM + Dynamic Feature + Dynamic Shape Coding | 1.24 |
| Stage- II | Verification Stage | 0.58 |
| Total runtime |  | 1.82 |

## 5. Conclusion

In this study, we have designed a framework to search query keyword across different scripts. The system, at first, takes help of on-the-fly script-wise keyword translation and then a line based word spotting approach using HMM was used to detect the query keyword. We proposed a novel unsupervised dynamic shape coding based scheme to group similar shape characters to reduce the similar shape character classes and thus improves the word-spotting performance. The reduced character set is obtained using dynamic shape coding which in turn improves recall performance. Next, that particular location of the text line image is verified with actual character-set of corresponding script. Thus improves the precision of word detection performance. To our knowledge dynamic shape coding has not been used for word spotting in earlier research work. To enrich the shape feature, the two-step word-spotting approach uses contextual information from the neighbour sliding windows.

The framework has been tested in multi-script framework consisting of three different scripts; Latin, Devanagari and Bengali to justify the efficiency. The experiment has been performed in low resolution scene images as well as video frames and we analysed the system performance with individual stages. From the comparative study we noted that our system outperforms the other existing systems. In future, the contrast or texture information can also be analysed to improve the text enhancement in scene/video images. The efficiency of two-stage dynamic shape coding based text detection approach can be taken forward for word spotting systems in other applications such historical documents, handwritten documents, etc. In future, we will try to integrate our framework with deep learning models where small set of dynamic shape codes can be used for word detection with limited data size.